\documentclass[letterpaper, 10 pt, conference]{ieeeconf}  

\IEEEoverridecommandlockouts                              

\usepackage{graphics} 
\usepackage{amsmath} 
\usepackage{amssymb}  
\usepackage{xcolor}
\usepackage{float}
\usepackage{graphicx}

\title{\LARGE \bf
Analysis of Interior Rubble Void Spaces at Champlain Towers South Collapse}

\author{Ananya Rao$^{1}$, Robin Murphy$^{2}$, David Merrick$^{3}$, and Howie Choset$^{1}$
\thanks{*Portions of this work were supported by NSF Grant CMMI-2140528.  Any opinions, findings, and conclusions or recommendations expressed in this material are those of the author(s) and do not necessarily reflect the views of the National Science Foundation.}
\thanks{$^{1}$A. Rao and H. Choset are with the Robotics Institute, Carnegie Mellon University, Pittsburgh, PA, USA ananyara@andrew.cmu.edu}%
\thanks{$^{2}$R. Murphy is with Texas A\&M University, College Station, TX, USA}%
\thanks{$^{3}$D. Merrick is with Florida State University, Tallahassee, FL, USA}%
}

\begin{document}

\maketitle
\thispagestyle{empty}
\pagestyle{empty}

\begin{abstract}
The 2021 Champlain Towers South Condominiums collapse in Surfside, Florida, resulted 98 deaths. 
Nine people are thought to have survived the initial collapse, and might have been rescued if rescue workers could have located them.
Perhaps, if rescue workers had been able to use robots to search the interior of the rubble pile, outcomes might have been better.
An improved understanding of the environment in which a robot would have to operate to be able to search the interior of a rubble pile would help roboticists develop better suited robotic platforms and control strategies. 
To this end, this work offers an approach to characterize and visualize the interior of a rubble pile and conduct a preliminary analysis of the occurrence of voids. 
Specifically, the analysis makes opportunistic use of four days of aerial imagery gathered from responders at Surfside to create a 3D volumetric aggregated model of the collapse in order to identify and characterize void spaces in the interior of the rubble. 
The preliminary results confirm expectations of small number and scale of these interior voids. 
The results can inform better selection and control of existing robots for disaster response, aid in determining the design specifications (specifically scale and form factor), and improve control of future robotic platforms developed for search operations in rubble.
\end{abstract}

\section{INTRODUCTION}
Improving the response to structural collapse disasters is a major focus of search and rescue robotics research. 
Due to the size and accessibility of areas to search, searching the interior of a rubble pile presents different challenges than searching the standing or partially damaged portions of a structure, and thus requires a different set of tools than what is conventionally available today.
We believe that robots should be in such a toolbox, and searching rubble piles necessitates different sizes and form factors of robots.
This work focuses on the collapse of the Champlain Towers South Condominium in June 2021, which is the third largest structural collapse in the United States~\cite{cbsSurfside}.
Of the 98 resulting deaths, nine were deemed to not be caused by crush injuries, indicating that these nine victims might have been rescued if they had been quickly found and extricated from the rubble~\cite{usaTodaySurfside, UKDailyMailSurfside}.

Rescue workers may be able to use uncrewed ground vehicles (UGVs) to search the interior of a rubble pile at a disaster site to help locate survivors, as robots can be sent into areas that cannot be accessed by human or canine rescue workers.
UGVs have been successfully used at the World Trade Center disaster~\cite{wtc}, and the Berkman Plaza II collapse~\cite{disasterRobotics, Berkman}, as well as in post-earthquake relief efforts in Mexico City~\cite{MexicoCity} and Christchurch~\cite{disasterRobotics}.
However, despite this, UGVs were not used for searching for survivors at Surfside.

One possible reason for the lack of UGVs is that UGVs are not designed to search the interior of rubble. 
Prior work \cite{disasterRobotics} has attempted to characterize the interior of rubble as a work envelope, but that characterization has been based on inferences from what robots to date have already encountered. 
Structural engineers have not been able to predict the occurrence of voids in ways that help with robot design specifications. 
Therefore, an improved understanding of the number, locations, and characteristics of void spaces in disaster sites is needed to inform the development of better suited robotic search and rescue platforms. 

\begin{figure*}
    \centering
    \includegraphics[width=\linewidth]{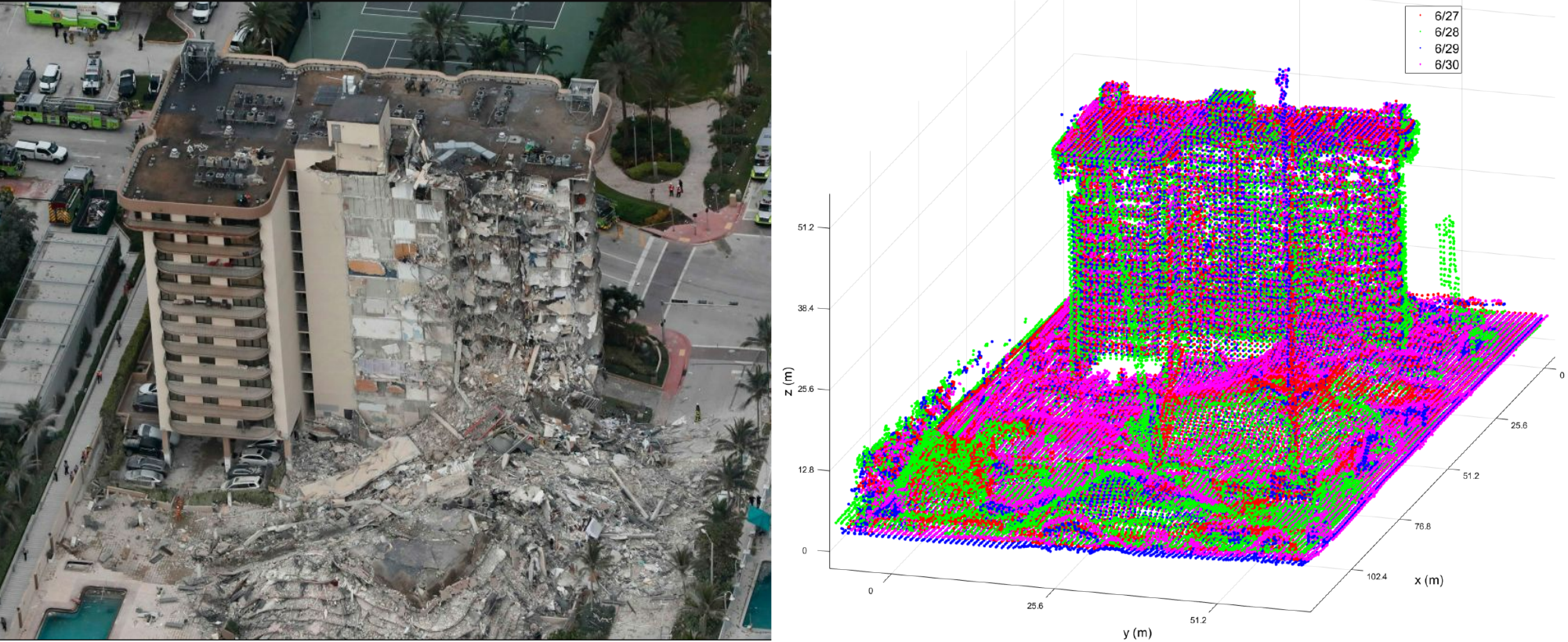}
    \caption{Aerial imagery collected to support the Surfside disaster response~\cite{uas} was processed to create an aggregated 3D model of the disaster site in the form of a point cloud stack (right), which was then analyzed to find and characterize voids. For contextualizing this 3D volumetric model, an aerial view of the Surfside collapse site~\cite{cnn} is shown (left).}
    \label{fig:surfside-overview}
    \vspace{-0.3cm}
\end{figure*}

This work provides a data-driven analysis of void spaces in the Surfside rubble, with the goal of characterizing and visualizing interior voids in order to inform future work.
The data used in this work was collected by flying an uncrewed aerial system (UAS) over the disaster site at Surfside in a grid-like flight pattern.
Aerial RGB and thermal images, and oblique imagery, were collected by responders for use in strategic planning on-site.
Multiple flights were conducted each day for the duration of the response, with the first flight occurring on June 25, 2021 at 16:30 local time.
The images collected were processed using Structure from Motion software (specifically Agisoft Metashape Professional) to create dense 3D point clouds of the disaster site (see Fig~\ref{fig:surfside-overview}). 
Agisoft Metashape was also used to generate orthophoto mosaics, which are raster images made by merging orthophotos (aerial or satellite photographs which have been transformed to correct for perspective so that they appear to have been taken from vertically above at an infinite distance) of the collapse.
This work focuses on data collected on four separate days (June 27-30, 2021) at similar times of day.
This limited subset of the available data is used as a proof of concept of using the approach presented in this work to locate and characterize void spaces.

In this work, the point clouds generated from data collected in four different UAS flights are used to identify and characterize voids in interior rubble, with the aim of better understanding the environments in which rescue UGVs must operate.
The constructed point clouds are manually aligned, and the resulting point cloud stack is systematically investigated to identify and locate potential void spaces. 
The potential voids and related oblique imagery are then evaluated to determine scale and cause of the voids. 
This approach allows roboticists to see voids in rubble, and offers evidence of their location, frequency of occurrence, and size.

The paper is organized as follows.
Prior work in disaster site analysis, the use of ground robots at other disaster responses, and void detection in 3D point clouds is discussed in Section~\ref{prior-work}. 
Our approach to locating and characterizing interior rubble voids is presented in Section~\ref{approach}.
Section~\ref{results} contains observations made, and analysis of identified voids.
Finally, conclusions and directions for future work are presented in Section~\ref{conclusion}.

\section{BACKGROUND AND PRIOR WORK}
\label{prior-work}
Prior work has attempted to characterize voids in rubble suitable for robotics, either through sensing with ground penetrating radar of simulated voids or from inference from cumulative studies; this work attempts to directly detect and characterize voids in an actual collapse using aerial imagery. Prior work has also considered detecting voids in 3D point clouds, however those efforts have focused on more structured point clouds of synthetic data with smaller number of points.  This work addresses large, complex and noisy 3D point clouds generated from real-world imagery.

\subsection{Analysis of Voids in Rubble}
Prior works have attempted to infer void characteristics from technical search tools, or to characterize \textit{in situ} voids with ground penetrating radar~\cite{hu}.
However, these efforts have been conducted in low fidelity physical simulations, and not with actual voids. 
Further, ground penetrating radar suffers performance degradation in heterogeneous conditions, due to signal scattering, thereby leading to limited applicability in the mixed-material environment of a rubble pile at a disaster site.

There is also analysis of building collapses from a civil engineering perspective.
For example, for the Surfside collapse, the nature and location of the rubble pile was analyzed in order to estimate the cause of the collapse, and some associated structural information~\cite{surfside}.
From this analysis, a simulation of the collapse can also be created.
However, such analyses do not consider the interior properties of the rubble pile itself, and do not provide information on what voids may have existed in the rubble.

The most comprehensive formal definition of the characteristics of rubble for robots appears in \textit{Disaster Robotics}~\cite{disasterRobotics} and related articles [20], [21].
The set of attributes presented in Disaster Robotics helps characterize the interior rubble as a robot work envelope, which is useful for selecting and deploying robots at a disaster site, as well as for developing new robotics rescue technologies. 
There are eight attributes: 1.scale, 2.verticality, 3.tortuosity, 4.surface properties, 5.severity of obstacles, 6.accessibility elements, 7.number of regions, and 8.non-navigational constraints, with 2-7 being collectively referred to as traversability. 
These characteristics overlap with, and go beyond, two other sets of attributes developed for ground robot navigation, see [18], [19].
This work focuses on determining scale and number of regions.

\subsection{Void Detection in 3D Point Clouds}
There exist several methods for detecting voids in point clouds.
One method has the user manually detect point cloud voids, and then is able to fill these point cloud voids in geometrically complex areas~\cite{yang}.
This approach however cannot process both smooth and sharp (or low and high frequency geometry) voids at the same time, as such properties are encoded into the parameters that are selected at run-time. 
Other methods create polygonal meshes from input point clouds using a Delaunay Triangulation algorithm, and then either perform a set of angle checks in order to determine which vertices are on the boundary of a void~\cite{dey}, or conduct a data search on the mesh to determine which vertices are present only in a single polygon, and thus lie on the void boundary~\cite{jun}.
Another work employs Ball Pivoting mesh generation~\cite{bird} to improve void detection.

While these methods have shown results in automatically detecting voids in 3D point clouds, they do not perform well on highly unstructured data, and point clouds that are missing sections, such as that of a rubble pile.
The approach presented in this work uses systematic manual inspection, as the focus is on characterizing the voids found, and not yet on automating void characterization.

\section{Locating and Characterizing Void Spaces}
\label{approach}
In this work, a systematic manual approach is used to find, locate and characterize void spaces in the rubble pile of the Surfside building collapse. 
The approach can be broadly divided into two high-level steps: locating potential voids spaces, and characterizing the identified voids. 

\subsection{Potential Void Space Identification}
\begin{figure}
    \centering
    \includegraphics[width=0.95\linewidth]{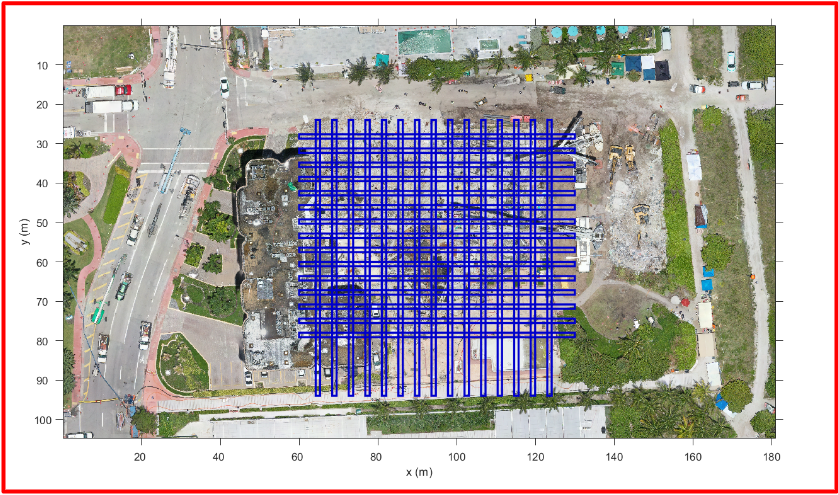}
    \caption{In order to systematically look for void spaces in the interior of the rubble pile, ``slices`` or cross sectional views are taken into the aggregated 3D volumetric model, or point cloud stack, to examine cross sections of the rubble pile on a 4m x 4m grid. The slices are 1m x 64m in the XY plane, and have the depth of the entire rubble pile. Here the slices are depicted on the orthophoto mosaic constructed from images gathered during a single UAS flight on June 27, 2021.}
    \label{fig:grid_slices}
    \vspace{-0.3cm}
\end{figure}

Finding potential void spaces involves processing the data gathered at Surfside, and then manually looking for voids in the following steps.
\begin{enumerate}
    \item Aerial images of the disaster site were collected from flying an uncrewed aerial system (UAS) to support response efforts~\cite{uas}. 
    \item Agisoft Metashape was used to process the collected data into dense 3D point clouds of the disaster site. 
    \item The 3D point clouds were manually aligned and ``stacked`` to create an aggregated 3D volumetric representation of the collapse (see Fig~\ref{fig:surfside-overview}).
    \item Cross sections were taken of the 3D volumetric model of the rubble pile to be able to look for potential voids. The cross sections were taken on a 4m x 4m grid, as depicted in Fig~\ref{fig:grid_slices}.
\end{enumerate}

The UAS flights over the Surfside site were conducted at several times each day during the response and recovery to map the disaster site~\cite{uas}. 
This analysis opportunistically uses the imagery from one mapping flight at roughly the same time each day for four days: June 27, 2021 13:30, June 28, 2021 13:30, June 29, 2021 13:00, and June 30, 2021 11:00 local time. 
These four days were chosen because it was a manageable set of data and it spanned days where the activity on the rubble pile was limited. 
Starting on July 1, 2021, operations increased and more people and equipment began to cover the surface of the rubble pile distorting the construction of point clouds. 


\subsection{Void Space Characterization}
Once potential void spaces have been identified, each one was characterized, specifically in terms of size and scale 
Additionally, each potential void space was analyzed to distinguish between naturally occurring voids and artifacts of excavation. 
The steps for analyzing each potential void space are as follows.
\begin{enumerate}
    \item The centroid location of the potential void was estimated, and then plotted on each of the four orthophoto mosaics of the Surfside site being considered in this work (June 27 - 30, 2021). The orthophoto mosaics were generated using Agisoft Metashape, with one orthophoto mosaic for each UAS flight.
    \item Dimensions of the potential void were estimated by inspecting the XZ plane and YZ plane cross sections of the void space. Here, the Z axis points up. 
    \item The estimated dimensions were used to approximate the size and volume of the potential void space.
    \item The orthophoto mosaics, and oblique imagery of the potential void, were inspected for excavation activity, and compared to void dimensions and shape seen in the cross sectional views in order to determine if the potential void occurred naturally or was created by excavation.
\end{enumerate}



\section{RESULTS AND DISCUSSION}
\label{results}
\begin{table*}
    \centering
    \begin{tabular}{|c|c|c|c|c|c|c|}
    \hline
        Void & Approximate & Maximum & Minimum & XZ Plane Cross & YZ Plane Cross & Cause \\
         & Volume (m$^3$) & Height (m) & Height (m) & Sectional Width (m) & Sectional Width (m) & \\
         \hline
         Yellow & 29.10 & 2.28 & 0.37 & 4.23 & 2.87 & Excavation \\
         Cyan & 34.53 & 1.88 & 0.33 & 4.25 & 6.25 & Excavation \\
         Orange & 41.71 & 1.45 & 0.15 & 5.03 & 6.91 & Excavation \\
         Purple & 28.71 & 1.19 & 0.20 & 6.37 & 5.06 & Excavation \\
         Pink & 10.94 & 1.14 & 0.26 & 3.41 & 5.73 & Naturally Formed \\
         Green & 10.33 & 0.82 & 0.23 & 2.69 & 5.91 & Naturally Formed \\
         \hline
    \end{tabular}
    \caption{Summary of characteristics of identified void spaces in the interior of the Surfside rubble pile.}
    \label{tab:summary}
    \vspace{-0.3cm}
\end{table*}

To frame our observations, the effort described in this paper focuses on characterizing voids, which will help establish the necessary form factor and perceptual constraints for search and rescue robots operating in the interior of rubble.
This work represents the first step: finding and confirming naturally occurring voids in the rubble pile.

\begin{figure*}
    \centering
    \includegraphics[width=0.9\linewidth]{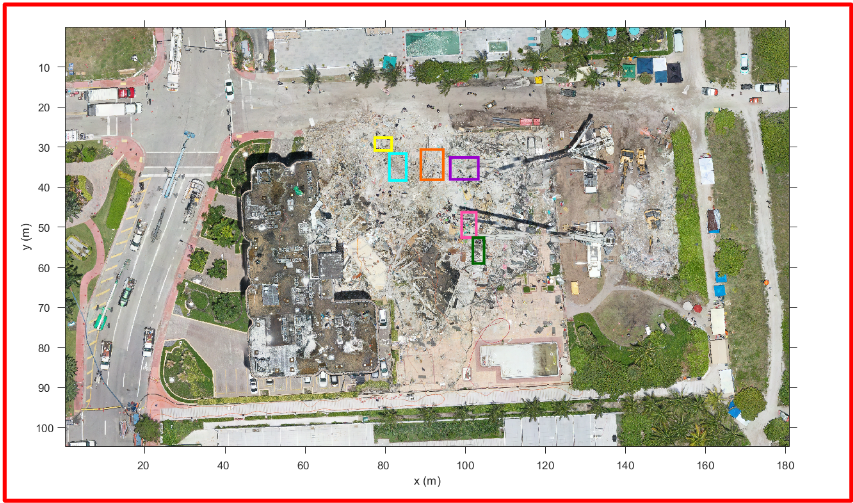}
    \caption{The locations and estimated XY plane bounding boxes of the six identified voids are overlaid on the orthophoto mosaic from June 27, 2021.}
    \label{fig:all_voids}
\end{figure*}

Consistent with hypotheses formed from \textit{Disaster Robotics}, Very few potential voids are seen in the rubble pile.
``Slicing`` the rubble pile model (i.e. inspecting cross sections of the 3D volumetric model) on a 4m x 4m grid produced only six candidate voids (depicted in Fig~\ref{fig:all_voids}) out of 225 possible locations. 
The candidate voids average a maximum height of 1.46m, a minimum height of 0.26m, and a cross sectional width of 4.89m.
A summary of their characteristics is presented in Table~\ref{tab:summary}.
While the voids themselves are not large enough to be survivable, they could offer breathing space, and show the types of void spaces that a robot would need to reach, inspect, and transit through.

\begin{figure*}
    \centering
    \includegraphics[width=\linewidth]{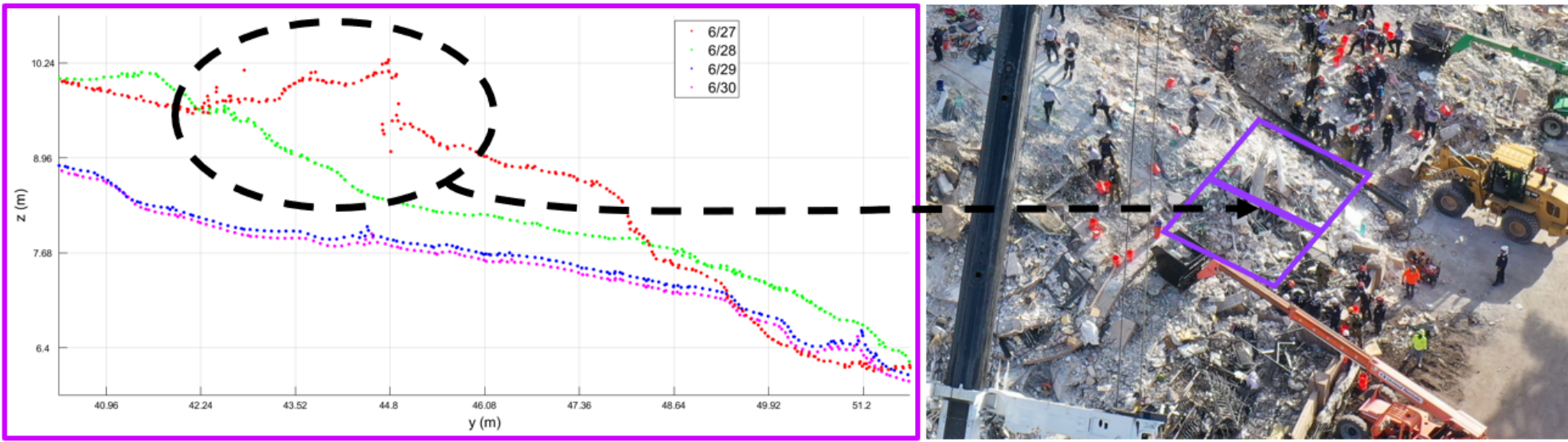}
    \caption{By inspecting oblique imagery of the purple void (right), and relating this to a cross sectional view of the purple void (left), it is seen that large objects and features are reasonably recognizable and consistent between the imagery and the layers of the constructed 3D volumetric model.}
    \label{fig:layer_valid}
\end{figure*}

The layers of the aggregated volumetric model constructed did appear to match the oblique imagery, implying that the layers are reasonably correct, and capture large objects and the ``stair steps`` of the layers of a pancake collapse.
An example of this is seen in Fig~\ref{fig:layer_valid}.

Our observations also support the hypothesis that robots deployed to similar disaster sites would need to be able to burrow into the rubble in order to be effective (this is consistent with \textit{Disaster Robotics}).
The voids seen in the Surfside collapse do not appear to form a connected set of tunnels.
The voids also did not appear to have entrances from the surface of the rubble.
A UGV would thus need to be able to burrow into the rubble to both access the voids, and also to traverse between them.

\begin{figure}
    \centering
    \includegraphics[width=\linewidth]{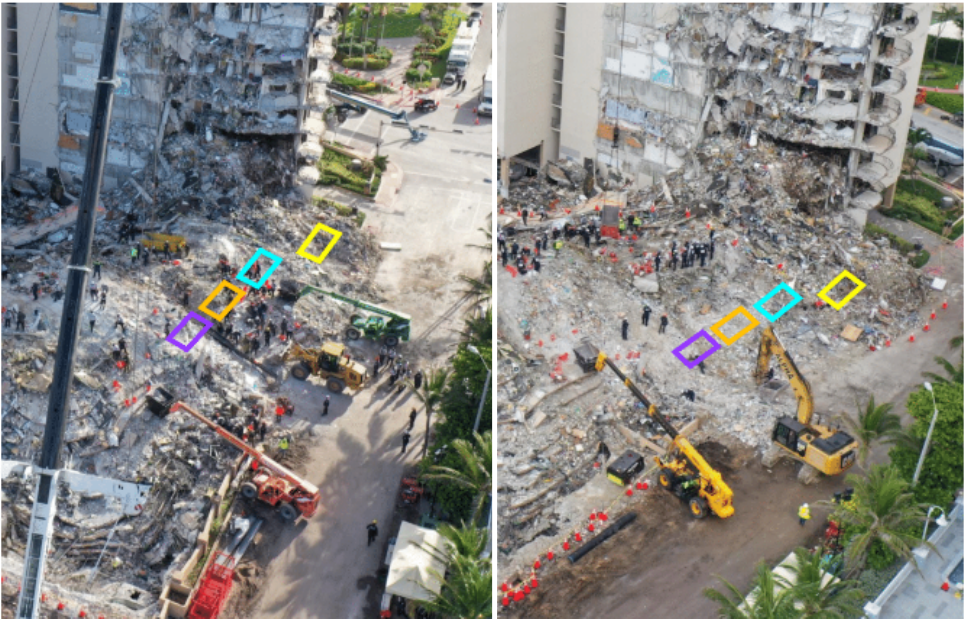}
    \caption{High levels of activity are seen in the oblique imagery from June 27, 2021 (left), and June 28, 2021 (right) in the region of the yellow, cyan, orange, and purple voids. It can be inferred from this that these potential voids are really artifacts of excavation.}
    \label{fig:excavation_voids}
    \vspace{-0.5cm}
\end{figure}

\begin{figure}
    \centering
    \includegraphics[width=\linewidth]{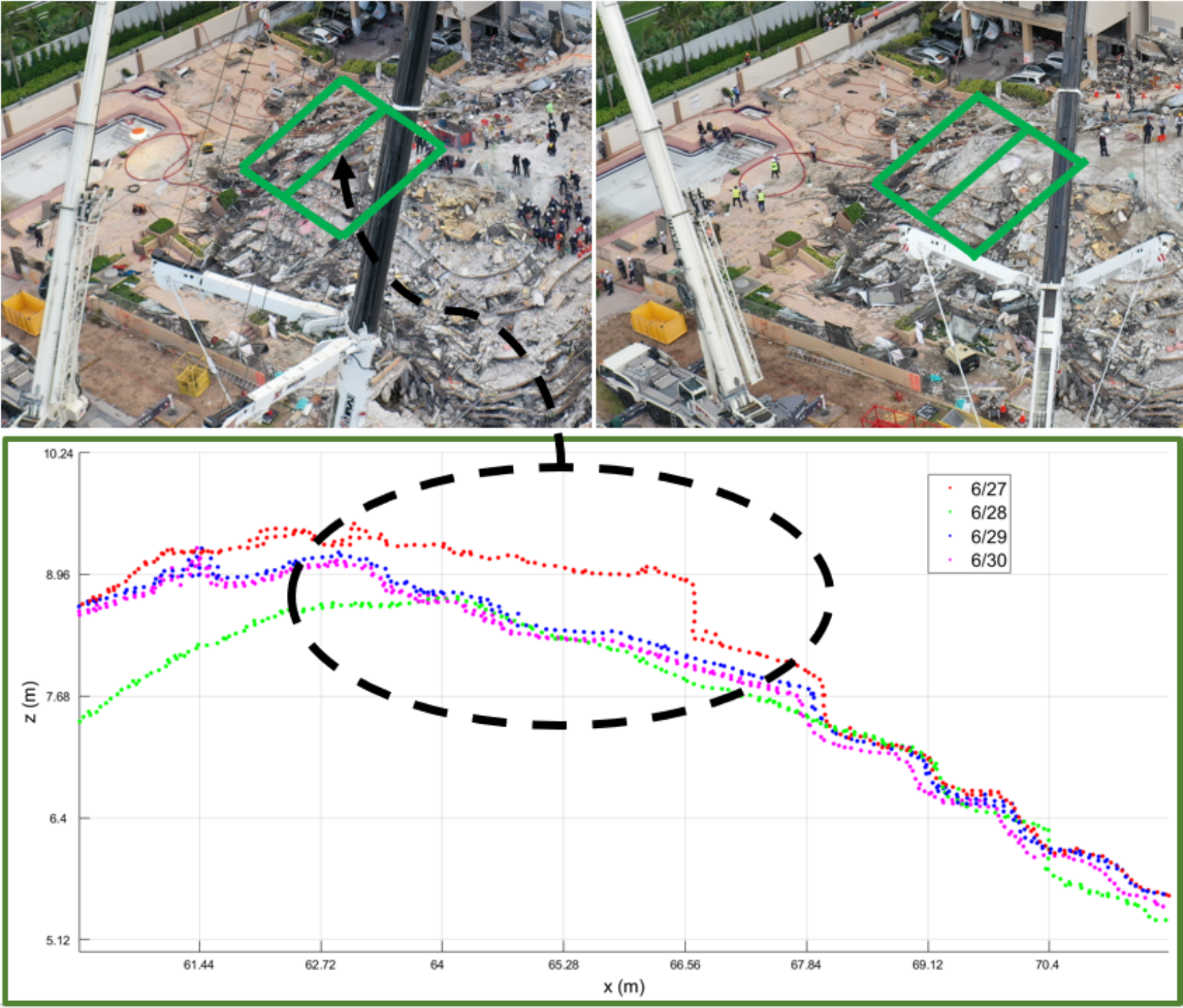}
    \caption{Some material was removed in the green void region between June 27, 2021 (top left) and June 28, 2021 (top right). There appears to be some slabs removed, which is consistent with the shape of the top of the void space seen in the XZ plane cross sectional view of the green void (bottom). This implies that the green void was naturally formed during the collapse.}
    \label{fig:green_obliques}
    \vspace{-0.8cm}
\end{figure}

Further, it is difficult to determine which voids are naturally occurring and which are artifacts of excavation. 
By examining oblique imagery from throughout the four days, four out of six of the candidate voids appeared to be the result of excavation (see Fig~\ref{fig:excavation_voids}).
Oblique imagery from June 27, 2021 and June 28, 2021 show high excavation activity in the regions of the yellow, cyan, orange, and purple potential voids.
This leads us to infer that the void space seen between the June 27 and June 28 layers is the result of excavation.

Two of the six voids, specifically the pink and green voids in Fig~\ref{fig:all_voids}, were in the pancake section of the collapse, and the associated oblique imagery suggests naturally forming voids were revealed by removing slabs.

A layer of material is visibly removed between the June 27th and June 28th (see Fig~\ref{fig:green_obliques}), however, the void seen between the corresponding layers of the digital surface model is larger than would be expected from excavation alone, based on the estimated thickness of the slabs removed.  
This implied that a naturally formed void space was underneath the layer of slabs that were removed.  

\begin{figure}
    \centering
    \includegraphics[width=\linewidth]{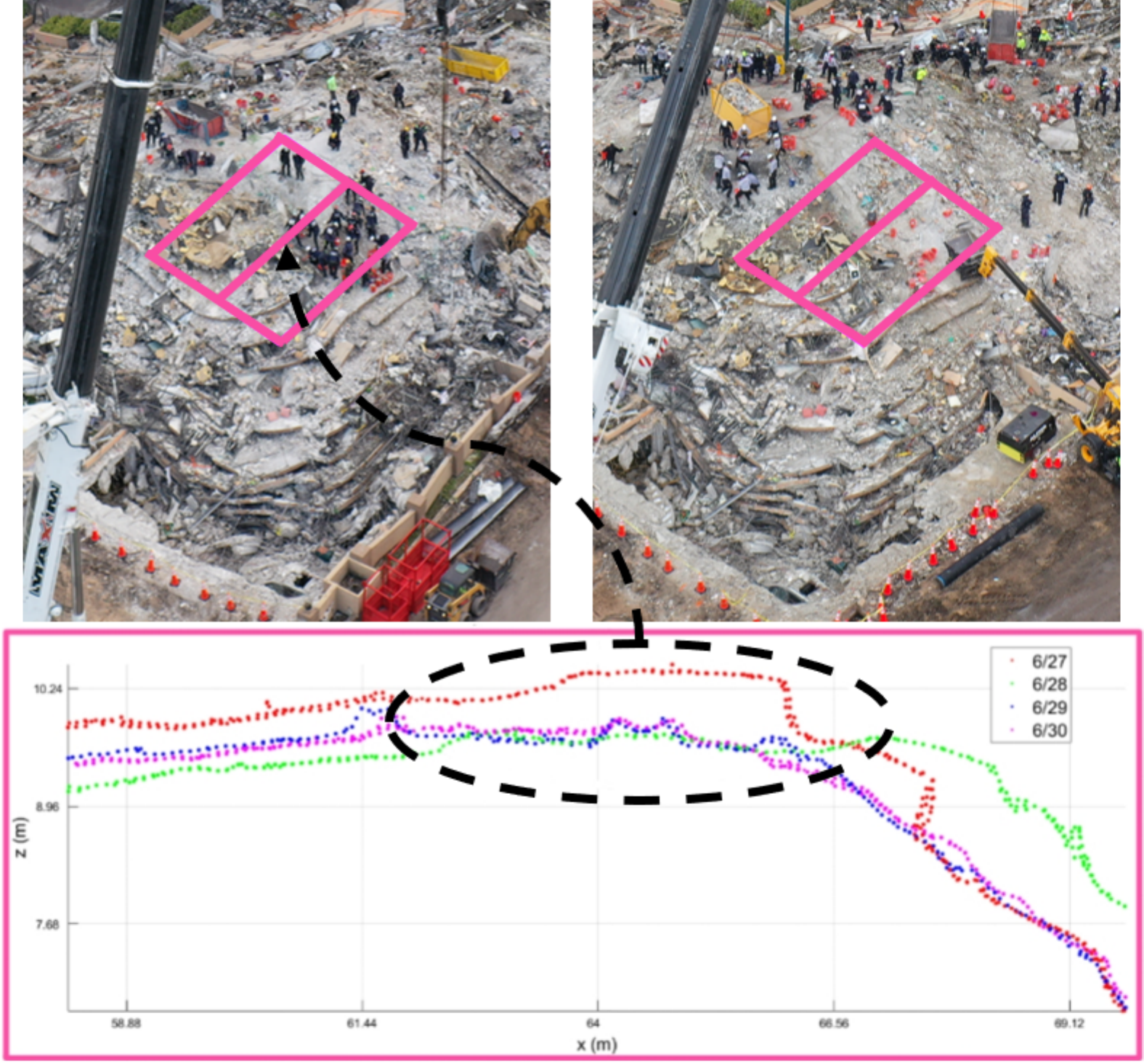}
    \caption{Some material was removed in the pink void region between June 27, 2021 (top left) and June 28, 2021 (top right). There appears to be some slabs removed, which is consistent with the shape of the top of the void space seen in the XZ plane cross sectional view of the pink void (bottom). This implies that the pink void was naturally formed during the collapse.}
    \label{fig:pink_obliques}
    \vspace{-0.3cm}
\end{figure}

Similarly, slabs were removed from the pink void region between June 27th and June 28th, with a void space that is larger than the estimated thickness of the slabs (see Fig~\ref{fig:pink_obliques}).
In both the green and pink voids, slabs act as a ``ceiling`` of sorts for the void space underneath.

Analysis of the voids shows that they may require robots with smaller form factors and more flexibility, which is consistent with predictions made in Disaster Robotics 2014.
The six voids varied in maximum height from 0.82m to 2.28m, minimum height from 0.15m to 0.37m, and cross sectional width from 2.69m to 6.91m.
The two voids in the pancake section (pink and green) were on the order of 0.25m - 1.14m in height. 
While not survivable, these voids do indicate the size constraints for a robot to be able to penetrate and navigate into the void, and search for bigger voids. 
They also provide insight into the form factors best suited for exploring these voids - smaller, more flexible robots, like snake robots, are best suited.

Candidate voids at Surfside did not show the verticality encountered in void spaces seen at the World Trade Center, parking garages, and mine disasters, however the depth of the rubble pile reinforces the notion that robots have to penetrate more than 6m into the rubble to reach areas where there might be survivable voids.

\section{CONCLUSIONS}
\label{conclusion}
This preliminary work establishes a promising approach to characterizing and visualising the interior of rubble that not only allows for formally characterizing voids as robot work envelopes, which can aid in the design of effective robots, but also offers data that could assist civil engineers in predicting where voids are likely to form.
Further, this work could help rescue workers focus search efforts on areas where voids are likely.
The approach enables void spaces in rubble to be visualized for the first time by using a 3D volumetric aggregated model of the entire collapse that can be examined as cross-sections or ``fly throughs."

However, further work is needed. The results showed an average translation error in point cloud alignment of about 80 cm, even though it is possible to detect interesting void spaces from point clouds, improvements would need to be made in order to more accurately characterize the dimensions of these voids.
Development of a tool to automatically register and align point clouds would help reduce the translation error.
Automatic point cloud alignment on this data is challenging due to it's complex and dynamic nature, and an analysis of existing methods for alignment supports the need for additional work in this space~\cite{bal}.
This work also only considers four days of data (27 June, 2021 - 30 June, 2021), which correspond to the ``top`` layers of the collapse. 
More work is needed to cover all of the days of data, but activity significantly increased on the site after 30 June 2021, so while voids may appear deeper in the rubble pile, it may be harder to distinguish between naturally occurring voids and artifacts of excavation. 

Further, this is an opportunistic \textit{post hoc} study; the aerial imagery was collected for strategic planning for the on-the-ground response and has notable gaps in frequency of data collection and viewpoints that would help ascertain voids. 

The data was also restricted to RGB images.
This approach can be improved both with dedicated data collection at higher frequency of UAS flights and from different angles, and also by using additional sensors, such as Lidar. 
More frequent mapping imagery and obliques, or the use of Lidar, would likely support automated registration of point cloud layers.
However, such data collection may interfere with the search and rescue priorities at a disaster site, and therefore, there is no guarantee that better data can be obtained, and most certainly, we do not want to interfere with the work flow of the rescue workers, as saving lives is the highest priority.

Despite these limitations, the preliminary results presented in this work confirm expectations of the small number and small scale of void spaces in rubble. 
These results can be used for better selection and control of existing robotic platforms for a disaster response. 
The purpose of this paper was to show that identifying the existence of voids is possible. 
The next step is then to provide in-depth analysis of the these voids, to both better characterize the rubble pile for possible locations of survivors, as well as understand how real rubble piles impact robot design. 
Such design considerations include size, number of degrees of freedom, payload, sensory needs (especially for confined spaces) and compliance in the mechanism. 
Compliance is important because the rubble pile is highly unstructured, and thus will have an abundance of unmodeled features. 
While this paper does not provide a detailed analysis of identified voids, further analysis will explore small size and actuation, structure or robot and payload, compliance due to lack of structure, and perception required to explore these voids.
The complexity of the rubble with the different types of collapse modes may favor a modular robotics approach in order to adapt to the different types of confined space environments in different collapse areas.

\addtolength{\textheight}{-12cm}   





\section*{ACKNOWLEDGMENT}

The authors thank Ananya Bal and Dr. Robert Ladig for their help with the data analysis and thoughtful discussions.






\bibliographystyle{unsrt}
\bibliography{references}

\begin{thebibliography}{10}

\bibitem{cbsSurfside}
K.~Barnett.
\newblock Surfside condo collapse is third largest building failure in
  country's history.
\newblock {\em CBS Miami}, 2021.
\newblock
  https://miami.cbslocal.com/2021/06/29/surfside-condo-collapse-third-largest-building-failure-us-history/.

\bibitem{usaTodaySurfside}
W.~Rhodes and R.~Ruiz-Goiriena.
\newblock At least 9 victims of 98 dead may have initially survived surfside
  condo collapse, but were not found by rescue teams, investigation shows.
\newblock {\em USA Today}, Aug. 24, 2021; Mar 14, 2022.
\newblock
  https://www.usatoday.com/story/news/nation/2021/08/24/surfside-condo-collapse-there-may-have-been-more-survivors-florida/8152538002/.

\bibitem{UKDailyMailSurfside}
S.~Thaler.
\newblock Nine of the 98 surfside victims may have survived the initial condo
  collapse: Fire rescue logs show woman was alive in rubble for ten hours.
\newblock {\em UK Daily Mail}, 2021.
\newblock
  https://www.dailymail.co.uk/news/article-9924083/At-9-98-Surfside-victims-SURVIVED-initial-condo-collapse.html.

\bibitem{wtc}
R.~R. Murphy.
\newblock Trial by fire.
\newblock {\em IEEE Robotics and Automation Magazine}, 11(3):50--61, 2004.

\bibitem{disasterRobotics}
R.~R. Murphy.
\newblock {\em Disaster robotics}.
\newblock MIT Press, 2014.

\bibitem{Berkman}
S.~Tadokoro, R.~Murphy, S.~Stover, W.~Brack, M.~Konyo, T.~Nishimura, and
  O.~Tanimoto.
\newblock Application of active scope camera to forensic investigation of
  construction accident.
\newblock In {\em 2009 IEEE Workshop on Advanced Robotics and its Social
  Impacts}, pages 47--50, 2009.

\bibitem{MexicoCity}
J.~Whitman, N.~Zevallos, M.~Travers, and H.~Choset.
\newblock Snake robot urban search after the 2017 mexico city earthquake.
\newblock In {\em 2018 IEEE International Symposium on Safety, Security, and
  Rescue Robotics (SSRR)}, pages 1--6, 2018.

\bibitem{uas}
R.~Murphy, D.~Merrick, J.~Adams, J.~Broder, A.~Bush, L.~Hart, and R.~Hawkins.
\newblock How robots helped out after the surfside condo collapse: Responders
  flew drones night and day to survey the collapse and search for survivors.
\newblock {\em IEEE Spectrum}, 2021.

\bibitem{cnn}
E.~C. McLaughlin.
\newblock What we know about the building that partially collapsed in surfside,
  florida.
\newblock {\em CNN}, 2021.
\newblock
  https://www.cnn.com/2021/06/24/us/what-to-know-about-the-surfside-building-collapse/index.html.

\bibitem{hu}
D.~Hu, S.~Li, J.~Chen, and V.~R. Kamat.
\newblock Detecting, locating, and characterizing voids in disaster rubble for
  search and rescue.
\newblock In {\em ” Advanced Engineering Informatics}, volume 111, pages
  109--124, 2019.

\bibitem{surfside}
X.~Lu, H.~Guan, H.~Sun, Y.~Li, Z.~Zheng, Y.~Fei, and L.~Zuo.
\newblock A preliminary analysis and discussion of the condominium building
  collapse in surfside, florida, us, june 24, 2021.
\newblock {\em Frontiers of Structural and Civil Engineering}, pages
  1097--1110, 2021.

\bibitem{yang}
L.~Yang, Q.~Yan, and C.~Xiao.
\newblock Shape-controllable geometry completion for point cloud models.
\newblock {\em The Visual Computer}, 33(3):385--398, 2017.

\bibitem{dey}
T.~K. Dey and J.~Giesen.
\newblock Detecting undersampling in surface reconstruction.
\newblock {\em Discrete and Computational Geometry}, 15:329--345, 2003.

\bibitem{jun}
Y.~Jun.
\newblock A piecewise hole filling algorithm in reverse engineering.
\newblock {\em Computer-Aided Design}, 37(2):263--270, 2005.

\bibitem{bird}
B.~Bird, B.~Lennox, S.~Watson, and T.~Wright.
\newblock Autonomous void detection and characterisation in point clouds and
  triangular meshes.
\newblock {\em International Journal of Computational Vision and Robotics},
  9(4):368 -- 386, August 2019.

\bibitem{bal}
A.~Bal, R.~Ladig, P.~Goyal, J.~Galeotti, H.~Choset, D.~Merrick, and R.~Murphy.
\newblock A comparison of point cloud registration techniques for on-site
  disaster data from the surfside structural collapse.
\newblock 2022.

\end{thebibliography}

\end{document}